\documentclass[letterpaper,10pt,conference]{ieeeconf}

\IEEEoverridecommandlockouts
\usepackage{amsmath,amssymb}
\usepackage{booktabs}
\usepackage{graphicx}
\usepackage{multirow}
\usepackage[hidelinks]{hyperref}
\usepackage{capt-of}
\usepackage{placeins}

\ifdefined\pdfobjcompresslevel\pdfobjcompresslevel=0\fi
\ifdefined\pdfcompresslevel\pdfcompresslevel=0\fi

\title{\LARGE\bf
A Reach-to-Push Curriculum for Learning Non-Prehensile Loco-Manipulation Across Object Families}

\author{Ebasa Temesgen, Minghao Zou, Maria Gini, and Karthik Desingh%
\thanks{The authors are with the Department of Computer Science and Engineering,
University of Minnesota, Minneapolis, MN, USA.
Correspondence: \texttt{temes021@umn.edu}.}}

\IEEEaftertitletext{%
\begin{center}
  \includegraphics[width=0.9\textwidth]{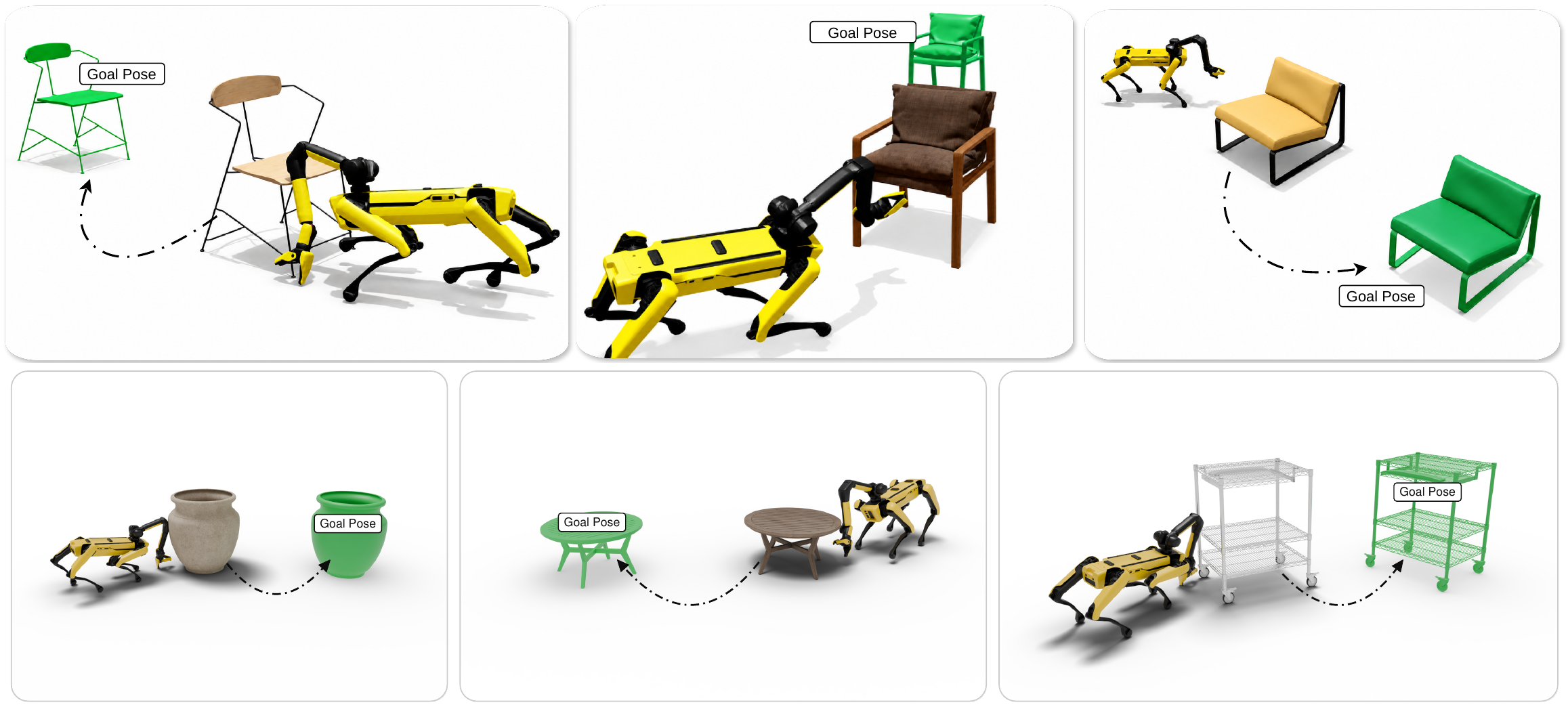}
  \captionof{figure}{\textbf{Pushing objects to a goal position and orientation.}
  The learned policy coordinates the base and arm of a legged mobile manipulator to push objects with varied geometry toward planar target poses (green).}
  \label{fig:teaser}
\end{center}
}

\begin{document}
\maketitle

\begin{abstract}
Arm-mediated pushing lets a mobile manipulator reposition large or
geometrically complex objects without first acquiring a grasp.
Learning this behavior is difficult because rewards based on
object motion provide little guidance for establishing a useful robot–object contact.
We present a hierarchical reinforcement-learning system
for goal-conditioned non-prehensile pushing with a quadrupedal
mobile manipulator.
A high-level policy jointly commands planar base velocity and six
arm-joint targets, while a frozen low-level locomotion policy stabilizes the legged base.
During training, we reward the robot for moving its end effector toward a sampled point near the object. As training progresses, we adjust the rewards to emphasize pushing the object to its goal.
In simulation, on 500 in-distribution test objects of each shape,
the policy achieves end-of-episode success rates of 77.0\% on
cuboids and 90.2\% on cylinders.
Success requires the object to remain upright and end within the goal position and orientation.
We also test on objects whose dimensions, mass, friction, and center-of-mass offsets lie beyond the training ranges.
The policy transfers to unseen combinations of mass and friction
whose sliding resistance remains within the training range,
but transport failures rise sharply beyond this range.
To test transfer beyond cuboids and cylinders, we train a separate
policy on one chair using simulator-provided object properties.
This policy also pushes seven unseen chairs and five table,
cart, and planter meshes to goal poses.
\emph{Project page:}
\href{https://reach-to-push.github.io/}{\texttt{https://reach-to-push.github.io}}

\end{abstract}

\begin{figure*}[t]
  \centering
  \includegraphics[width=0.91\textwidth]{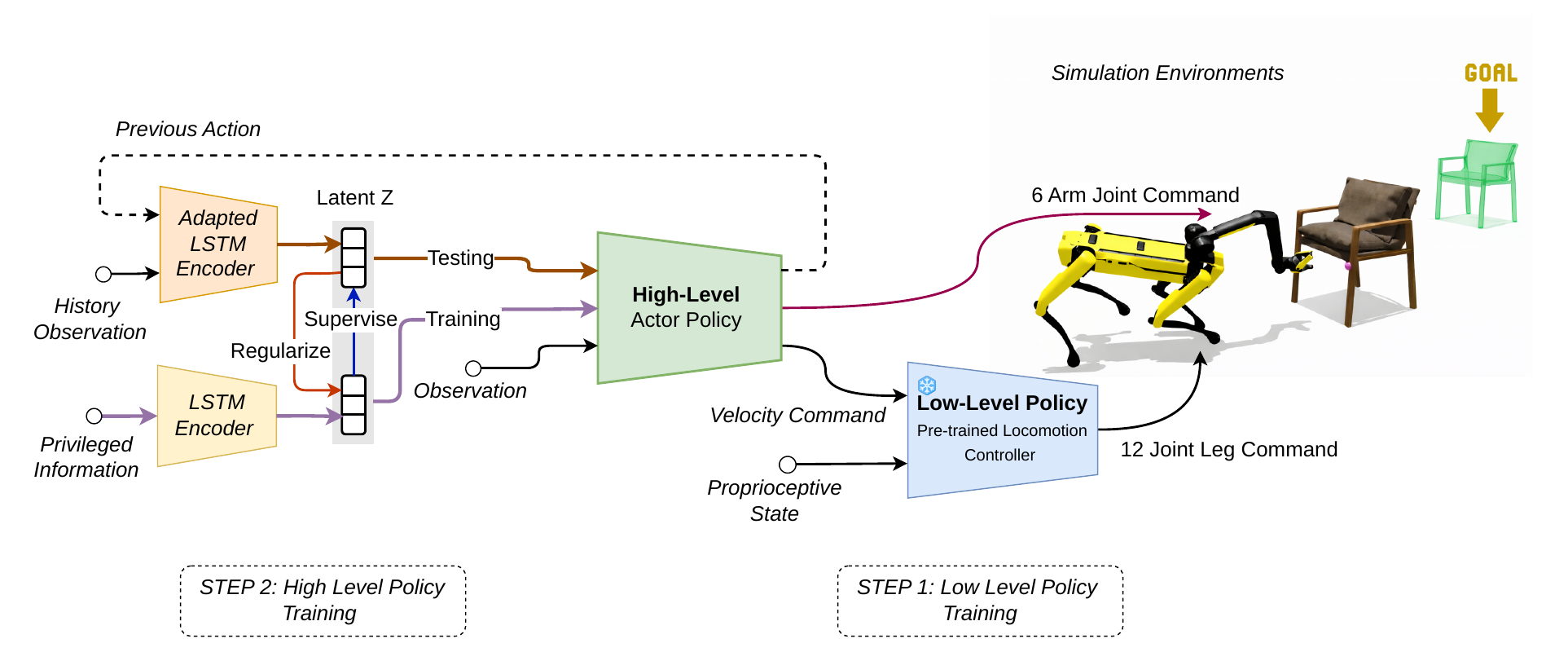}
  \caption{\textbf{Hierarchical policy architecture.}
  The high-level actor receives task state and an adaptation latent and outputs planar base velocity plus six arm-joint targets.
 A frozen locomotion policy maps the base command and proprioception to twelve leg-joint targets. Separate joint controllers track the arm targets.
   During training, the privileged LSTM encodes simulator-provided
object information and supervises the adapted LSTM.
At test time, the adapted LSTM processes observation history
and supplies the latent vector to the high-level policy.}
  \label{fig:system}

\end{figure*}

\begin{figure}[t]
  \centering
  \includegraphics[width=\columnwidth]{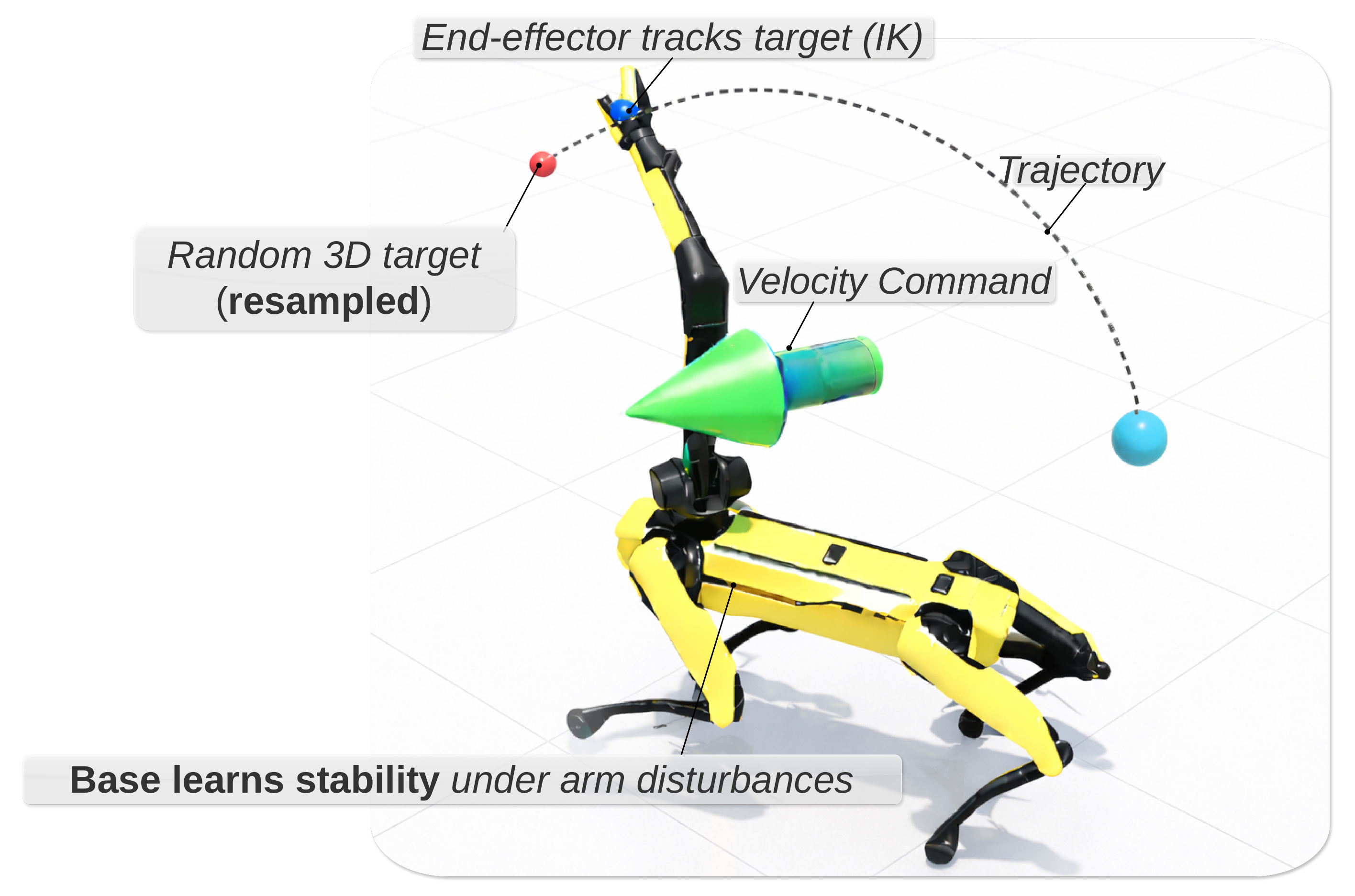}
  \caption{\textbf{Locomotion-prior pretraining.}
  The arm tracks resampled 3-D targets while the base follows velocity commands, exposing the leg controller to varied arm configurations and upper-body disturbances.}
  \label{fig:prior}

\end{figure}

\section{Introduction}

Robots in homes, hospitals, warehouses, and laboratories must often reposition objects that are too bulky, heavy, or irregular to grasp, such as chairs, carts, bins, and planters.
Non-prehensile pushing avoids the grasp and extends the usable workspace of a mobile robot, but its outcome depends strongly on the robot--object contact location, object geometry, friction, and mass \cite{stuber2020let,mason1986mechanics,lynch1996stable}.
For a legged mobile manipulator, the task is also coupled to balance. In arm-mediated pushing, the arm establishes contact to move
the object while the legs reposition and stabilize the robot's
floating base.

Recent learned controllers demonstrate pushing and loco-manipulation on quadrupeds \cite{Jeon2023WholeBodyQuadruped,dadiotis2025dynamic,liu2024visual} and show the value of separating task decisions from locomotion stabilization \cite{huang2024hilmares,2025relic}.
This separation does not by itself solve contact acquisition.
A policy trained only from distant pose rewards rarely discovers useful end-effector interactions, and a target sampled from one hand-selected face does not extend naturally from boxes to chairs, tables, or carts.
The central question in this work is therefore: \emph{how can reinforcement learning discover useful arm contacts across object families without hand-labeled contact locations for each object?}

We use \emph{object family} for objects that share a shape class, such as cuboids, cylinders, or chairs, while their size and physical properties vary. Unlike convex objects such as cuboids and cylinders, many everyday
objects, including chairs and carts, have non-convex shapes with
recesses or gaps.
To test transfer across chair geometries and object types,
we train on one chair and evaluate on previously unseen chairs
and other objects.

Our approach uses a reward curriculum that initially emphasizes
reaching the object. The reaching reward provides useful feedback before the policy learns to move the object.
At the start of each episode, we sample a target point from a set of candidate reaching locations defined relative to the object.
A reaching reward encourages the end effector to approach
this target.
The target is used only in the reward and is not observed by
the policy.
A separate stability reward discourages object toppling.
As training progresses, we reduce the weight of the reaching reward
and introduce a completion reward, activated only when the object
is at the goal.
This shifts the training emphasis from establishing contact
to pushing the object to its goal pose.

This paper makes three contributions:
(1) a reaching-reward curriculum for contact acquisition that samples targets either on the faces of a fixed box around the object or on a cached point cloud of the object mesh, without per-object contact annotations;
(2) a hierarchical arm-mediated pushing policy that combines a frozen disturbance-aware locomotion prior with a settling stage that teaches it to stop once the object is placed; and
(3) an arm-mediated pushing study covering parametric
in-distribution and out-of-distribution (OOD) conditions, twelve
unseen everyday-object meshes, and contact-specific failure
analysis.

\section{Related Work}

% \textbf{Non-prehensile pushing.}
Classical work relates applied contact, friction, and support
geometry to planar object motion
\cite{mason1986mechanics,lynch1996stable}.
Learning-based methods reduce reliance on exact mechanics and
have transferred tabletop non-prehensile skills from simulation
to physical systems \cite{lowrey2018nonprehensile}.
Mobile manipulation adds continuous base repositioning, so the robot can reach more contact locations around the object.
Prior systems plan feedback-controlled pushes for carts and
passive-caster objects
\cite{Scholz2011CartPushing,Mericli2013PushManipulationCasters},
while recent model-free constrained RL learns dynamic goal
pushing with mobile manipulators \cite{dadiotis2025dynamic}.
Hierarchical whole-body quadrupedal pushing has also been
explored across variations in object geometry and physical
properties \cite{temesgen2026wholebody}.

Learning arm-mediated pushing requires discovering contacts that
move the object toward its goal.
Effective contact locations vary across object geometries, and
rewards based on object motion provide little guidance before
contact is established.
For non-prehensile loco-manipulation, prior work guides exploration
using contact candidates generated by a grasping algorithm and
a separate exploration critic \cite{tolomei2026contact}.
Our approach uses sampled reaching targets defined relative to the object and uses a reward curriculum to reduce reaching guidance
as training progresses.

% \textbf{Legged loco-manipulation and adaptation.}
Hierarchical policies enable quadrupeds to manipulate objects
while retaining robust locomotion
\cite{Jeon2023WholeBodyQuadruped,huang2024hilmares,2025relic}.
Visual loco-manipulation policies and large-scale randomized
training further broaden the reachable task and object
distributions \cite{liu2024visual,Zhou2025AllInOne}.
Rapid and regularized adaptation infer latent dynamics from
recent interaction history
\cite{kumar2021rma,fu2022deepwholebodycontrollearning}.
These advances improve locomotion robustness and manipulation
under varying conditions, but they do not by themselves resolve
the exploration difficulty of acquiring useful arm contacts.
In particular, goal-conditioned pushing requires the robot
to coordinate base repositioning and end-effector approach
before object-pose rewards become informative.

We build on established hierarchical control principles,
using a frozen locomotion prior to stabilize the base while
a high-level policy coordinates base and arm motion.
Our contribution lies in sampled reaching targets, defined on a fixed box or on a cached object point cloud, and a staged curriculum that shifts learning from contact acquisition toward goal-directed object transport.

\section{Method}

\subsection{Task and Hierarchical Controller}

Let the initial planar object pose be $(p_o,\psi_o)$ and the commanded goal pose be $(p_g,\psi_g)$.
The objective is to bring the object to the goal without toppling it or making unintended contact with the robot body.
The high-level policy acts through
\begin{equation}
 a_t=[v_x,\,v_y,\,\omega_z,\,\phi,\,\theta,\,h,\,q^{\mathrm{arm}}_{1:6}]\in\mathbb{R}^{12},
 \label{eq:action}
\end{equation}
where the first three entries are body-frame planar velocity commands, $(\phi,\theta,h)$ are torso roll, pitch, and height commands, and the remaining entries are arm-joint position targets, while the gripper remains fixed.
The pretrained locomotion policy~\cite{2025relic} is kept frozen
and converts base commands and proprioception into twelve
leg-joint targets, while arm impedance controllers track
$q^{\mathrm{arm}}_{1:6}$.

The locomotion prior is pretrained while the arm tracks randomly resampled end-effector targets (Fig.~\ref{fig:prior}).
This exposes the leg controller to a broad distribution of upper-body configurations and disturbances before task learning.
The frozen locomotion policy stabilizes the base while the
high-level policy learns to coordinate base motion and all six
arm joints for pushing.

The state-based actor observes base linear and angular velocity, projected gravity, its previous action, arm positions and velocities, end-effector-to-object displacement, object-to-goal displacement, and relative object and goal rotations.
The sampled target is used only to compute the reaching reward
and is not provided as an input to the policy.
Object poses are obtained directly from the simulator.
We therefore study pushing control assuming known object poses,
without estimating them from sensor data.

\subsection{Policy Optimization}\label{sec:policy-opt}

The high-level actor is a $512$--$256$--$128$ MLP optimized with PPO \cite{schulman2017proximalpolicyoptimizationalgorithms} at 50\,Hz.
The implementation also contains a teacher--student LSTM adaptation module \cite{fu2022deepwholebodycontrollearning}. In the trained parametric policy, its latent converged to a near-constant value, and replacing the student latent with the privileged teacher latent changes in-distribution success by at most 1.6 points.
We therefore treat the evaluated controller as a feedforward policy on the current state and attribute no result to adaptation.

\subsection{Contact-Target Acquisition}

At the start of training, rewards based on object motion
provide little guidance before the robot makes contact.
We therefore define an interaction set $\mathcal S_{\mathrm{int}}$
around the object and uniformly sample a target at each
episode reset:
\begin{equation}
 p_c\sim\mathcal U(\mathcal S_{\mathrm{int}}),\qquad
 p_c^w=T_{wo}p_c .
\end{equation}
Here, $p_c$ is expressed in the object frame, and $T_{wo}$
maps it to world coordinates.

\begin{figure}[t]
  \centering
  \includegraphics[width=0.86\columnwidth]{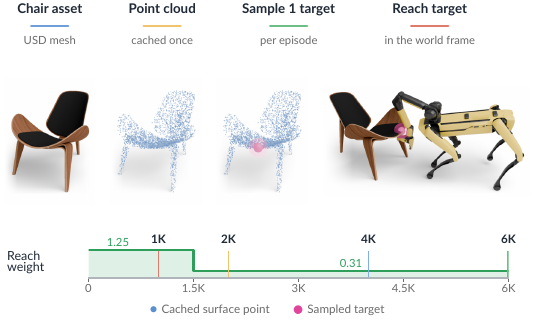}
  \caption{\textbf{Mesh-surface contact sampling.}
  A cached surface point cloud is filtered to obtain candidate
  contact points, and one target is sampled per episode.
  % The evaluated policies use the box-face targets described
  % in the text.
  }
  \label{fig:sampling}

\end{figure}

For the policy trained on cuboids and cylinders, $\mathcal S_{\mathrm{int}}$ consists
of the four vertical faces of a $0.7\times0.7\times1.0$\,m box
centered in the object frame, offset 3\, cm inward.
The chair policy uses the same box in its initial training. In its settling stage, we instead sample targets from a cached point cloud of the chair mesh, filtered for reachability and clearance (Fig. \ref{fig:sampling}).

% No face is labeled as the correct pushing face.

The target remains fixed relative to the object throughout
the episode.
It is used only to compute the reaching reward.
The policy does not observe the target.
It can move away from the sampled target as the curriculum
places greater emphasis on moving the object to the goal. 

% A mesh-surface sampling variant filters a cached point cloud
% for reachability and clearance (Fig.~\ref{fig:sampling}).

% This variant was not enabled for the policies evaluated here.

\subsection{Reward and Curriculum}

The shaped objective combines contact acquisition, object-pose regulation, progress, stability, and a contact penalty:
\begin{align}
 r_{\mathrm{reach}}&=\exp(-\lVert p_{ee}-p_c^w\rVert_2/\sigma_r^2),\\
 r_{\mathrm{pose}}&=\exp(-\lVert K_o-K_g\rVert_2/\sigma_p^2),
\end{align}
where $K_o$ and $K_g$ stack the eight bounding-box corners of the object and goal.
Additional terms reward object velocity aligned with the goal direction, reward staying upright, and penalize unintended non-arm contacts (Table~\ref{tab:reward}).
Episodes terminate when the object tilts beyond $45.6^\circ$ ($\cos\theta<0.7$) or the simulation becomes unstable.

\begin{table}[t]
\centering
\caption{\textbf{Arm-mediated task reward.} Arrows denote curriculum changes. The last four terms are added in the settling stage.}
\label{tab:reward}
\scriptsize
\renewcommand{\arraystretch}{1.08}
\resizebox{\columnwidth}{!}{%
\begin{tabular}{lccc}
\toprule
\textbf{Term} & \textbf{Form} & $\boldsymbol{w}$ & $\boldsymbol{\sigma}$ \\
\midrule
Pose alignment & $\exp(-\lVert K_o-K_g\rVert_2/\sigma^2)$ & 4.00 & 0.77 \\
Reach contact & $\exp(-\lVert p_{ee}-p_c^w\rVert_2/\sigma^2)$ & $2.50{\to}0.63$ & 0.59 \\
Velocity to goal & $\exp(\hat v_o^\top\hat d_g/\sigma^2-1)$ & 0.16 & 0.71 \\
Object stability & $[1+\exp(500(\theta-20^\circ))]^{-1}$ & 0.40 & -- \\
Sparse completion & $\mathbf 1(e_{xy}\leq5\text{ cm}\wedge e_\psi\leq5^\circ)$ & $0{\to}4.00$ & -- \\
Undesired contact & \#non-EE links with force $>0.2$\,N & $-0.05$ & -- \\
\midrule
\multicolumn{4}{l}{\textit{Settling stage, active while the object is within C2:}} \\
Object at rest & $\exp(-\lVert v_o\rVert_2/\sigma)$ & 2.00 & 0.05 \\
Arm released & $\mathbf 1(\text{no arm--object contact})$ & 1.00 & -- \\
Base at rest & $\exp(-\lVert a_{\mathrm{base}}\rVert_2^2/\sigma^2)$ & 1.00 & 3.32 \\
Object speed & $\lVert v_o\rVert_2$ within 20\,cm of the goal & $-1.00$ & -- \\
\bottomrule
\end{tabular}%
}

\end{table}

Early training emphasizes reaching with $w_{\mathrm{reach}}=2.5$.
After 2.8K iterations, this weight decreases to 0.63. After 4.8K iterations, the sparse completion weight increases from 0 to 4.0.
This schedule creates three behavioral phases: discover contact, learn sustained transport, and refine final pose.
One high-level policy is trained through all three phases.

\begin{figure*}[t]
  \centering
  \includegraphics[width=0.9\textwidth]{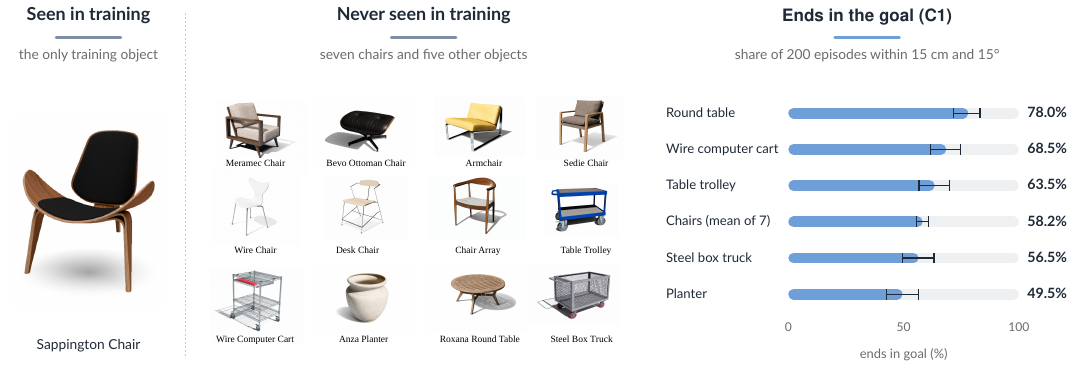}
  \caption{\textbf{Chair policy on unseen objects.}
  Left: the Sappington chair, the only training object. Middle: seven chairs and five other objects never seen in training. Right: share of episodes in which the settled chair policy leaves the object within the goal pose, with 200 episodes per object and 1,400 for the chair mean. Error bars show 95\% confidence intervals.}
\label{fig:unseen}
\end{figure*}

% \begin{figure}[t]
%   \centering
%   \includegraphics[width=\columnwidth]{figures/contact_shift_fixed.pdf}
%   \caption{\textbf{Contact over training (chair-trained policy).}
%   (A) One episode at four checkpoints; dots mark contacts (blue: arm, orange: legs/body).
%   (B) Contacts of all 64 episodes on the chair and around it seen from above, goal straight up; dashed arc: rear region.
%   Numbers: share of contact by arm only and legs/body only, and of arm contacts at the rear.}
%   \label{fig:contact-evolution}
%   \vspace{-0.7em}
% \end{figure}

\subsection{Settling Stage}\label{sec:settling}

The trained policy often continues pushing after the object
reaches the goal, moving it away again.
We add a settling stage to teach the robot to stop pushing
and leave the object at the goal. The simulator limits the policy's base velocity, torso pose,
and arm joint commands before passing them to the controllers.
This is called action clipping \cite{fujita2018clipped}.
The policy often requests values far beyond these limits,
so small changes during training still produce the same
clipped commands.
This makes it difficult to learn how to slow down and stop.

Before PPO fine-tuning, we train the high-level policy to predict
its previously recorded clipped actions from the corresponding
observations.
This adjusts the policy's outputs so that small changes during
further training can affect the commands sent to the controllers.
We then continue training the high-level policy with PPO and
the additional settling rewards in Table~\ref{tab:reward}.

\section{Experiments}

\subsection{Protocol and Metrics}

We train the high-level policy in Isaac Lab
\cite{Baginski2025IsaacLab} with 4096 parallel environments
for 22.4K PPO iterations on 50 cuboids and 50 cylinders.
We then continue training with the settling procedure
described in Sec.~\ref{sec:settling}.
Each episode lasts at most 30\,s in a $5.9\,\mathrm{m}\times5.9\,\mathrm{m}$ arena, with a random goal yaw and the robot starting 2.0--3.2\,m behind the object.
Each OOD condition moves one factor outside its training range (Table~\ref{tab:domain-rand}), and the combined-physics condition shifts mass, friction, and CoM together. Every condition and family is evaluated on 500 freshly sampled variants with fixed seeds.

\begin{table}[t]
\centering
\caption{\textbf{Train and OOD randomization ranges.}}
\label{tab:domain-rand}
\scriptsize
\setlength{\tabcolsep}{10.0pt}
\renewcommand{\arraystretch}{1.08}
\begin{tabular}{lcc}
\toprule
\textbf{Factor} & \textbf{Train} & \textbf{OOD} \\
\midrule
Mass (kg) & $[2,8]$ & $[8,18]$ \\
Static friction & $[0.6,1.2]$ & $[0.1,0.6]$ \\
Dynamic friction & $0.8\times$ static & $0.8\times$ static \\
CoM offset, $x,y$ (cm) & $0$ & $[-10,10]$ \\
Width/length (m) & $[0.25,0.75]$ & $[0.25,1.25]^\dagger$ \\
Height (m) & $[0.4,1.0]$ & $[0.4,1.8]^\dagger$ \\
Goal distance (m) & $[1.5,2.5]$ & $[1.5,2.5]$ \\
\midrule
\multicolumn{3}{l}{\scriptsize $^\dagger$At least one dimension lies outside the training range.} \\
\bottomrule
\end{tabular}

\end{table}

Criterion C1 uses 15\,cm and $15^\circ$, and the stricter C2 uses 10\,cm and $10^\circ$. Note that cylinders are scored on position only. An episode succeeds if the object ends the 30\,s episode inside the tolerance and has not toppled.
We also report the median terminal position error $e_{\mathrm{pos}}$ and wrapped yaw error $e_\psi$.

\begin{table}[t]
\centering
\caption{\textbf{Arm-mediated pushing over parametric object families.}  Share of 500 objects per condition that end the episode upright and within criterion C1 (15\,cm, $15^\circ$) or C2 (10\,cm, $10^\circ$), with median final errors. }
\label{tab:main}
\scriptsize
\setlength{\tabcolsep}{3.6pt}
\renewcommand{\arraystretch}{1.08}
\resizebox{\columnwidth}{!}{%
\begin{tabular}{lrrrr@{\hspace{8pt}}rrr}
\toprule
 & \multicolumn{4}{c}{\textbf{Cuboid}} & \multicolumn{3}{c}{\textbf{Cylinder}} \\
\cmidrule(lr){2-5}\cmidrule(lr){6-8}
\textbf{Condition} & \textbf{C1} & \textbf{C2} & $\boldsymbol{e_{\mathrm{pos}}}$ & $\boldsymbol{e_\psi}$ & \textbf{C1} & \textbf{C2} & $\boldsymbol{e_{\mathrm{pos}}}$ \\
 & \textbf{(\%)} & \textbf{(\%)} & \textbf{(cm)} & \textbf{(deg)} & \textbf{(\%)} & \textbf{(\%)} & \textbf{(cm)} \\
\midrule
% rows generated by figure_scripts/build_table3_ends.py from learn2push/results/finish_rc_test
In-distribution & 82.6 & 77.0 & 4.2 & 2.1 & 90.8 & 90.2 & 2.8 \\
\midrule
Geometry & 67.2 & 57.2 & 7.9 & 2.9 & 82.6 & 77.4 & 4.5 \\
Mass & 39.6 & 36.4 & 26.6 & 12.7 & 43.4 & 41.4 & 39.8 \\
Friction & 78.6 & 69.0 & 5.1 & 3.1 & 94.2 & 88.6 & 3.6 \\
CoM offset & 75.2 & 72.4 & 4.2 & 2.4 & 74.2 & 74.0 & 3.2 \\
Combined physics & 74.6 & 69.2 & 4.4 & 2.9 & 90.4 & 87.0 & 3.3 \\
\bottomrule
\end{tabular}
}

\end{table}

\subsection{Parametric Performance and Distribution Shift}\label{sec:param}

In distribution, 77.0\% of cuboids and 90.2\% of cylinders
end within C2 (Table~\ref{tab:main}). Out of distribution, \emph{mass} alone gives the lowest C2
success, at 36.4\% on cuboids and 41.4\% on cylinders.
This condition increases object mass while keeping friction
within the training range. We also test increased mass, reduced friction, and a center-of-mass offset together.
This \emph{combined-physics} condition uses the same higher
mass range as the \emph{mass} condition.
Its C2 success rates are higher, at 69.2\% on cuboids and
87.0\% on cylinders.

The mass increase makes the objects harder to slide, while
the friction reduction makes them easier to slide by reducing
the required pushing force.
These changes therefore partly offset one another
(Sec.~\ref{sec:failures}).
Changing three properties in this combined test does not
make it harder than changing mass alone.

In the \emph{geometry} condition, we test cuboids and cylinders
with dimensions outside the training ranges.
These changes affect both object stability and how the object
turns when pushed.
Objects that are tall relative to their base width are more
likely to topple.
For wider cuboids, the policy must also adjust where and how
it pushes to achieve the goal orientation. C2 success decreases from 77.0\% to 57.2\% on cuboids and
from 90.2\% to 77.4\% on cylinders.

Before the settling stage (Sec.~\ref{sec:settling}), the robot
often continued pushing after the object reached the goal,
moving it away before the episode ended.
With the object at the goal, settling reduced the percentage
of control steps with the base command at its velocity limit
from 89--100\% to at most 3.5\%.
The proportion of episodes in which the object entered C2
and then left it also fell from 9.7\% to 4.3\%.
Together, these improvements raised end-of-episode C2 success
across all 6{,}000 test episodes from 60.6\% to 70.0\%.  The additional training helped the policy leave the object
at the goal, but increased toppling under geometry shifts.
Toppling rates rose from 13.6\% to 17.0\% on cuboids and
from 10.6\% to 14.0\% on cylinders.

% \begin{figure}[t]
%   \centering
%   \includegraphics[width=0.85\columnwidth]{figures/am_rollout_trajectory.pdf}
%   \caption{\textbf{Representative successful AM rollout} (top view). The base arcs around the object while the arm maintains and redirects contact.}
%   \label{fig:am-rollout}

% \end{figure}

\begin{figure*}[t]
  \centering
  \includegraphics[width=0.92\textwidth]{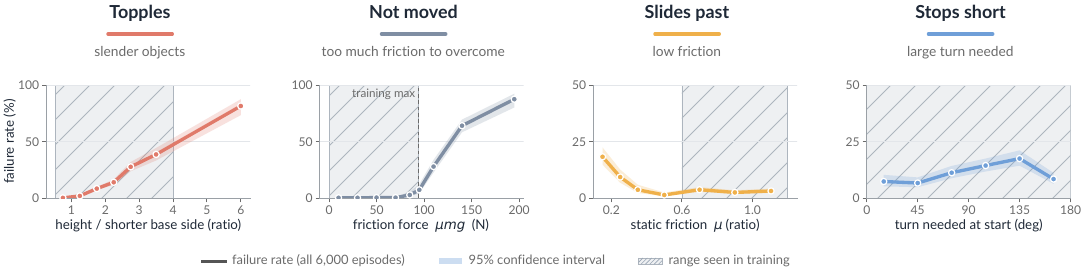}
  \caption{\textbf{Each failure mode follows one object property.}
  Rate of each failure mode over the 6{,}000 episodes of Table~\ref{tab:main} (12 conditions $\times$ 500), binned by the property that drives it. Bands show 95\% confidence intervals, and grey shading shows the range seen in training.}
  \label{fig:failure-drivers}
\end{figure*}

\subsection{Transfer Beyond Parametric Families}\label{sec:transfer}

We test whether a policy trained on one chair can push objects
with different structures (Fig.~\ref{fig:unseen}).
None of the twelve meshes was used for training.
After settling, 58.2\% of episodes on the seven unseen chairs
end within C1, compared with 36.4\% before settling.
Success rates increase on all twelve objects, although the
post-settling estimate for one chair remains within its
pre-settling 95\% confidence interval.
For the wire cart, success rises from 20.5\% to 68.5\%. Across all twelve objects, the proportion of episodes that
reach C1 at any point increases from 47.0\% to 68.0\%.
The policy also leaves objects at the goal more reliably,
as the gap between reaching C1 and ending within C1 narrows
from 12.5 to 7.8 percentage points. The learned strategy transfers to tables, carts, and planters without object-specific retraining.
% However, success varies across objects, and these twelve
% meshes do not establish transfer to arbitrary objects.
% The chair policy observes object properties that a real robot would have to estimate. We therefore run the same test with the policy trained on cuboids and cylinders, which does not observe them and has never seen a mesh. It ends 17.5\% of chair episodes and 49.8\% of episodes on the other five objects within C1, compared with 58.2\% and 63.2\% for the chair policy.

% We also record which robot links touch the object. The policy trained on cuboids and cylinders makes 83.8\% of its contact with the gripper on those objects. On the unseen objects, the gripper and other arm links together account for 81.8\% of its contact. For the chair policy on unseen objects, the gripper accounts for 17.0\% of contact time, the other arm links for 30.5\%, and the legs and body for 52.5\%.

% \subsection{Qualitative AM Behavior}

% In Fig.~\ref{fig:am-rollout} the base does not drive the object; it repositions so that the arm keeps a contact from which a push moves the object toward the goal, while the arm sets the direction of motion.
% Locomotion creates contact opportunities and the arm exploits them, a coupling that a fixed approach-then-push sequence lacks.

\begin{figure}[t]
  \centering
  \includegraphics[width=0.85\columnwidth]{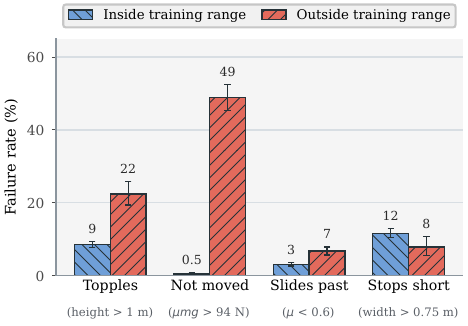}
  \caption{\textbf{Three of four failure modes concentrate outside the training range.}
Failure rates inside and outside the training range for each property (condition shown below each group). Error bars show 95\% confidence intervals.}
  \label{fig:failures}
\end{figure}

\subsection{Ablations}

Table~\ref{tab:ablation} compares the full method with variants
that remove the reaching reward, sparse-reward curriculum,
or the adaptation encoder.
The comparison uses training on one chair and evaluation
on held-out chairs.

\begin{table}[t]
\centering
\caption{\textbf{Ablations on held-out chairs.} Training chair (left), tested on the Meramec chair. Share of successful episodes (right).}
\label{tab:ablation}
\begin{minipage}[c]{0.26\columnwidth}
  \centering
  \includegraphics[width=0.9\linewidth]{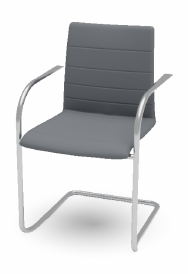}
\end{minipage}\hfill
\begin{minipage}[c]{0.71\columnwidth}
  \centering
  \scriptsize
  \setlength{\tabcolsep}{4pt}
  \renewcommand{\arraystretch}{1.15}
  \begin{tabular}{lrr}
  \toprule
  \textbf{Variant} & \textbf{C1 (\%)} & \textbf{C2 (\%)} \\
  \midrule
  No sampled reach reward & 0.0 & 0.0 \\
  No sparse-reward curriculum & 6.3 & 0.0 \\
  No adaptation encoder & 39.0 & 36.9 \\
  Full method & 69.0 & 67.6 \\
  \midrule
  Privileged-input expert & 79.9 & 69.5 \\
  \bottomrule
  \end{tabular}
\end{minipage}
\vspace{-0.8em}
\end{table}

% \begin{table}[t]
% \centering
% \caption{\textbf{Ablation results on held-out chairs.}
% Policies are trained on one chair and evaluated on chairs
% not used for training.}
% \label{tab:ablation}
% \scriptsize
% \setlength{\tabcolsep}{3.1pt}
% \renewcommand{\arraystretch}{1.12}
% \begin{tabular}{lrrrr}
% \toprule
% \textbf{Variant} & \textbf{C1} & \textbf{C2}
% & $e_{\mathrm{pos}}$ & $e_\psi$ \\
%  & \textbf{(\%)} & \textbf{(\%)}
%  & \textbf{(cm)} & \textbf{(deg)} \\
% \midrule
% No sampled reach reward & 0.00 & 0.00 & -- & -- \\
% No sparse-reward curriculum & 6.25 & 0.00 & 184.46 & 77.89 \\
% No adaptation encoder & 39.00 & 36.91 & 51.19 & 44.17 \\
% \method{} & 69.03 & 67.64 & 29.70 & 18.75 \\
% Privileged-input expert & 79.86 & 69.51 & 12.37 & 13.10 \\
% \bottomrule
% \end{tabular}

% \end{table}

The largest drops in reported success occur when the reaching
reward or sparse-reward curriculum is removed.
Without the reaching reward, success is 0\% under both criteria.
Without the curriculum, C1 success is 6.25\%, while C2 success
remains 0\%. Removing the adaptation encoder also reduces reported C2
success, from 67.64\% to 36.91\%.
The privileged-input expert reaches 69.51\% at C2,
slightly above the full method.

\subsection{Failure-Mode Analysis}\label{sec:failures}

The failures explain the generalization results above.
We label each failed episode of Table~\ref{tab:main} at C2 by how it ends: the object \emph{topples}, is \emph{not moved} (never within 50\,cm of the goal), \emph{slides past} (enters the 10\,cm tolerance and leaves it), or \emph{stops short} (otherwise).
These modes account for 93\% of all failures, and each tracks one property of the object (Fig.~\ref{fig:failure-drivers}).

To slide an object, the robot must overcome its friction force $\mu m g$.
The policy moves almost every object below 90\,N (0.5\% failures) but fails on 64--88\% above 120\,N, and the transition lies at the largest friction force in the training distribution, about 94\,N (8\,kg at $\mu{=}1.2$).
This explains the mass result of Sec.~\ref{sec:param}. In the mass variations, 70\% of objects exceed 94\,N, whereas low friction keeps 98\% of the equally heavy objects of the combined-physics condition below it.
The boundary of the forces seen in training becomes a boundary of capability. A push applied above the base creates a moment about the bottom edge, whereas the restoring moment from gravity grows only with the base width, so the contact that slides a squat box rotates a slender one. Toppling rises from 0.2\% for objects no taller than they are wide to 82\% beyond four times.

Friction is also what brings an object to rest, so on slippery objects small corrective pushes near the goal carry it out again. After settling, 3.1\% of objects in the training friction range leave the goal once reached, but 18\% of those with $\mu{<}0.2$. Stopping short grows with the rotation the goal requires, from 7\% below 30$^\circ$ to 16\% for 90--150$^\circ$. One contact produces both translation and rotation, so a large turn must be corrected while the object moves, and these objects come close (median 21\,cm) without satisfying both tolerances. Toppling (2.6$\times$), sliding past (2.2$\times$), and not
moving (0.5\% inside versus 49\% outside) concentrate outside
the training ranges (Fig.~\ref{fig:failures}).
Stopping short remains common within the training ranges.

These findings suggest that training and evaluation should
consider combinations of physical properties, alongside their
individual ranges. Mass and friction jointly determine sliding
resistance, while height relative to base width affects toppling
risk. An object can therefore fall outside an individual training
range without requiring a greater pushing force. Friction also creates a tradeoff between transport and placement.
Lower friction helps the robot move heavier objects but makes
stopping at the goal harder. Evaluations should therefore
distinguish reaching the goal from leaving the object there
at the end of the episode.

Stopping short within the training distribution suggests that
broader domain randomization alone may not resolve final-pose
errors. The higher failure rate on larger turns motivates
investigating how policies coordinate translation, rotation,
and contact repositioning. One direction is to learn to predict
how a push changes the object's position and orientation,
using recent interactions to estimate its dynamics.
Such predictions could help the policy select contacts and
actions that correct position and orientation together.

\section{Conclusion}

We trained a hierarchical policy to push objects to a goal pose using a reaching reward toward a target sampled from candidate locations defined relative to the object and a curriculum that shifts the emphasis toward goal placement.
In simulation, this policy ended 77.0\% of cuboids and 90.2\% of cylinders within 10\,cm and $10^\circ$, which shows that a coarse target without contact labels or a mesh is sufficient to learn pushing on these families.
We observed that the trained policy kept pushing after the object arrived, so we added a settling stage. Mean success rose from 60.6\% to 70.0\%, with more toppling under geometry shifts, which shows that stopping must be trained explicitly.

We then tested the chair policy on twelve unseen objects. It reached 58.2\% on chairs, but it observes object properties directly, so reliably pushing chairs without those inputs remains a challenge. The failure analysis reveals a trade-off between transport and placement. High sliding resistance makes objects difficult to move, while lower friction helps transport but increases the risk of sliding past the goal. Tall objects with a narrow base are also more likely to topple. Stopping short persists within the training ranges, especially when the goal requires a large turn. A push that corrects orientation can also change position, so the policy struggles to coordinate these corrections near the goal.

\section*{Acknowledgments}
The authors thank Xun Tu for his contributions to the writing and
experiments of an earlier version of this work, and for helpful
discussions during its early stages.

\bibliographystyle{IEEEtran}
\bibliography{refs}

\end{document}